# DEEP NEURAL NETWORK FOR PIER SCOUR PREDICTION

MAHESH PAL
Professor, Department of Civil Engineering,
National Institute of Technology Kurukshetra,
136119, India, Email: mpce_pal@yahoo.co.uk, mahesh.pal@nitkkr.ac.in

**Abstract:** With the advancement in computing power over last decades, deep neural networks (DNN), consisting of two or more hidden layers with large number of nodes, are being suggested as an alternate to commonly used single-hidden-layer neural networks (ANN). DNN are found to be flexible models with a very large number of parameters, thus making them capable of modelling very complex and highly nonlinear relationships existing between inputs and outputs.

This paper investigates the potential of a DNN consisting of 3 hidden layers (100, 80 and 50 nodes) to predict the local scour around bridge piers using field data. To update the weights and bias of DNN, an adaptive learning rate optimization algorithm was used. The dataset consists of 232 pier scour measurements, out of which a total of 154 data were used to train whereas remaining 78 data to test the created model. A correlation coefficient value of 0.957 (root mean square error = 0.306m) was achieved by DNN in comparison to 0.938 (0.388m) by ANN, indicating an improved performance by DNN for scour depth perdition. Encouraging performance on the used dataset in the work suggests the need of more studies on the use of DNN for various civil engineering applications.

## 1. INTRODUCTION

Scour around bridge piers occurs due to changes in flow pattern around it because of partial blocking of stream flow. Blockage of flow due to bridge pier creates adverse pressure gradient on its upstream causing boundary layer to go under a three-dimension separation (Bateni et al. 2007). This separation of flow around the bridge pier leads to a change in the shear stress distribution around it resulting in excess sediment removal near the structure (Kothyari et al. 2002). Removal of excessive material leads to reduction in bed elevation near the piers, thus exposing the foundations of a bridge. The temporal variation in scour and the depth of scour near the bridge pier depend on the characteristics of flow, pier and river-bed material. The physical process of pier scour is a complicated process which makes it difficult to develop a methodology for scour prediction.

Several methods/formulas are proposed to estimate the equilibrium depth of local scour near the bridge piers. Being collected from small-scale laboratory experiments with non-cohesive and cohesive uniform bed material under steady-flow conditions is a major drawback of these formulas (Kandasamy and Melville 1998) and found performing poorly when used with the dataset acquired from field conditions (Jones 1984). Possible reasons of their poor performance may also be attributed to over simplification or ignoring the complexities of natural rivers by assuming uniform flow, using a constant flow depth with non-cohesive bed materials (Mueller and Wagner 2005).

Soft computing techniques, like back-propagation neural network (BPNN) are extensively used in predicting scour around bridge pier and abutments (Kambekar and Deo 2003; Azmathullah et al. 2006; Bateni et al. 2007; Lee et al. 2007; Kaya 2010; Ebtehaj et al. 2017; Eghbalzadeh et al. 2018; Hosseini et al. 2018). Comparison of predicted scour with various empirical equations suggests an improved performance by the BPNN approach.
Within last few years, several studies reported the use of deep neural network (DNN) in several civil engineering problems and found them performing well in comparison to the existing modeling approaches (Zhang et al. 2018; Zhou and Chen 2018; Deng et al. 2018; Kumar and Abraham 2019; Dick et al. 2019; Ding et al. 2019; Nguyen et.al. 2019). Keeping in view the improved performance of DNN based regression models, this paper explore its potential in predicting the scour near bridge piers using field dataset. To compare the performance of proposed DNN, a BPNN was used.

## 2. DEEP NEURAL NETWORK (DNN)

The basic element of a BPNN is the processing node. Each processing node behaves like a biological neuron and performs two functions. First, it sums the values of its inputs. This sum is then passed through an activation function to generate an output. Any differentiable function can be used as an activation function, $f$. All the processing nodes in BPNN are arranged into layers, each fully interconnected to the following layer. There is no interconnection between the nodes of the same layer. In a BPNN, generally, there is an input layer that works as a distribution structure of the data being inputted to the network and not used for any type of processing. After this layer, one or more processing layers called the hidden layers follows. The final processing layer is called the output layer. A neural network with two or more hidden layers having large number of nodes and using sophisticated mathematical modelling is generally called deep neural network.

All the interconnections between each node have an associated weight. When a value is passed from the input layer, down these interconnections, these values are multiplied by the associated weight and summed to derive the net input ($n_j$) to the unit

$$n_j = \sum_i w_{ji} o_i$$

where $w_{ji}$ is the weight of the interconnection to unit $j$ from unit $i$ (called input ) and $o_i$ is the output of the unit $i$. The net input obtained by the above equation is then transformed by the activation function to produce an output ($o_j$) for the unit $j$.

Traditionally, sigmoid and hyperbolic tangent are two extensively used nonlinear activation functions with a BPNN. Activation functions introduce non-linearity in the neural network so as to learn more complex features present in the data. Saturation and sensitivity to changes around their mid-point were found to be two major problems with the sigmoid and hyperbolic tangent functions (Goodfellow et al. 2016).

The rectified linear activation function (RELU; Nair and Hinton 2010) is a piecewise linear function and considered to be a major algorithmic change within last decades for the design of deep neural network (Goodfellow et al. 2016). RELU is one of the most popularly used activation functions in the deep learning which output the input value itself if it is positive, otherwise output would be zero. It is easier to train and found to achieve better performance than other activation functions with DNN. The RELU function is defined as:

$$f(n_j) = max(0, n_j)$$

Random weight initialisation is normally used with a standard BPNN because of the reason that stochastic gradient descent approach uses randomness in order to find optimal set of weights for the specific mapping function from inputs to outputs with the given dataset. Initializing BPNN with the correct weights is an important factor for proper functioning of neural network. The weights selected before the start of training the network should also be in a reasonable range.

Xavier weight initialization (Glorot and Bengio 2010) was proposed as the weights initialization technique for DNN because of the poor performance of random weight initialization with standard gradient descent based optimisation approach. This approach assigns the weights from a Gaussian distribution with zero mean and some finite variance, thus allowing the variance of the outputs of a layer to be equal to the variance of its inputs.

Learning rate is also an important user-defined parameter used to adjust the weights of the BPNN. Most of the studies reporting the application of back-propagation neural network in civil engineering used learning rates values which were randomly set by the user (a value between 0 and 1) based on the past experiences and earlier reported works, suggesting it's often hard to get its correct values. Before the introduction of adaptive learning rate methods, the gradient descent algorithms used with BPNN were updating the network weights with the help of a learning rate, the objective function and its gradient. To improve the working of traditional gradient descent algorithms, adaptive gradient descent (using adaptive learning rate) algorithms, which could adaptively tune the learning throughout the training process was proposed and used with DNN (Goodfellow et al. 2016). In this study, adaptive moment estimation (Adam; Kingma and Ba 2015) based optimization algorithm was used to update network weights during training in place of the classical stochastic gradient descent method. Adam computes individual learning rates for different parameters and requires setting several user-defined parameters. In this study, default values of all user-defined parameters as suggested by Kingma and Ba (2015) were used and found working well with this data.

Similar to simple BPNN, deep neural network required setting of several user defined parameters. These parameters includes, number and type of hidden layers, nodes in each hidden layer, activation function for output, hidden and dropout layers, weight initialization method, optimization algorithm, updaters (i.e. learning rate optimization algorithm), batch size (i.e. number of training samples used in one iteration) and number of epochs (i.e. one epoch is defined as when an entire training dataset has passed once through the neural network both in forward and backward direction).

## 3. DATA SET AND METHODOLOGY

Out of the total 493 field data for pier scour measurements (Mueller and Wagner 2005), a total of 232 measurements of upstream scour were selected and used in this study. This dataset was divided randomly in a way such that 154 data were used for training purposes and the remaining 78 data to test the models (Pal et al. 2014). Seven input parameters namely pier shape factor (Ps), pier width (Pw), skew of the pier to approach flow (skew), velocity of the flow (V), depth of flow (h), D50 (i.e. the grain size of bed material in mm for which 50 percent is finer) and gradation of bed material ($\sigma$) were used to predict the scour depth. Properties of training and test datasets used in the study are provided in Table 1. Keeping in view of better performance with raw data than non-dimensioned dataset by different machine learning algorithms, this study uses the dimensioned dataset only. For further details about this data readers are referred to Pal et al. (2014).

To compare the performance of DNN and BPNN modeling approaches, three parameters namely correlation coefficient, root mean square error (RMSE) and mean absolute error values calculated with testing dataset were used. The predictive performance of both DNN and BPNN depends on the choice of optimal value of several user-defined parameters. Extensive trials were carried out to find out the optimal values of different user-defined parameters by comparing the RMSE values with testing dataset with both neural networks. Table 2 provides optimal values of user-defined parameters for the dataset used in this study.

## 4. RESULTS

Correlation coefficient (CC), root mean square error (RMSE) and mean absolute error (MAE) values obtained with testing dataset using BPNN and deep neural networks are provided in Table 3. Results from Table 3 indicate improved performance by DNN in terms of all three parameters in comparison to BPNN algorithm. The Figure 1 provide the plot between actual and predicted pier scour depth achieved by using DNN with dropout layers and BPNN with the test dataset, suggesting an improved performance by DNN.

**Table 1.** Characteristics of the train and test data used in this study

| Input parameter | Train data | | | | Test data | | | |
|---|---|---|---|---|---|---|---|---|
| | Min | Max | Mean | St. dev. | Min | Max | Mean | St. dev. |
| $P_s$ | 0.70 | 1.30 | 0.97 | 0.21 | 0.70 | 1.30 | 0.99 | 0.20 |
| $P_w$ | 0.30 | 5.50 | 1.56 | 1.16 | 0.30 | 5.50 | 1.40 | 1.15 |
| *skew* | 0 | 85 | 9.26 | 18.63 | 0 | 65 | 9.90 | 18.37 |
| V | 0.20 | 4.50 | 1.64 | 0.89 | 0 | 3.20 | 1.30 | 0.68 |
| h | 0.30 | 22.50 | 4.55 | 4.02 | 0 | 22.40 | 3.80 | 3.58 |
| $D_{50}$ | 0.12 | 95 | 18.98 | 26.76 | 0.15 | 95 | 19.47 | 25.10 |
| $\sigma$ | 1.20 | 20.30 | 3.65 | 3.29 | 1.20 | 21.80 | 3.61 | 2.90 |
| Scour | 0.10 | 7.10 | 1.12 | 1.27 | 0.10 | 6.20 | 0.94 | 1.06 |

The unit of measurements for $P_w$, $h$ and scour depth is in meter, velocity of flow is in meter/second, $D_{50}$ is in mm and *skew* is measured in degrees.
$P_s$ = 1.3 for square nosed-piers, 1.0 for round-nosed piers and 0.7 for sharp-nosed piers.

**Table 2.** Optimal value of user-defined parameters DNN and BPNN

| Algorithm used | User-defined parameters |
|---|---|
| Deep neural network | Three Hidden layers (100, 80, 50 nodes), Activation function ReLU, Weight initiation-XAVIER, Batch size=5, Updater=Adam, Epochs=15000 |
| Back-propagation neural network | Learning rate =0.2, momentum =0.1, hidden nodes =8, hidden layer=1, number of iterations =1500 |

**Table 3.** Correlation coefficient, RMSE and MAE values with test dataset

| Modelling approach | RMSE (m) | Mean Absolute error (MAE) | Correlation Coefficient (CC) |
|---|---|---|---|
| Back propagation neural network | 0.390 | 0.297 | 0.937 |
| Deep neural network | 0.306 | 0.227 | 0.957 |

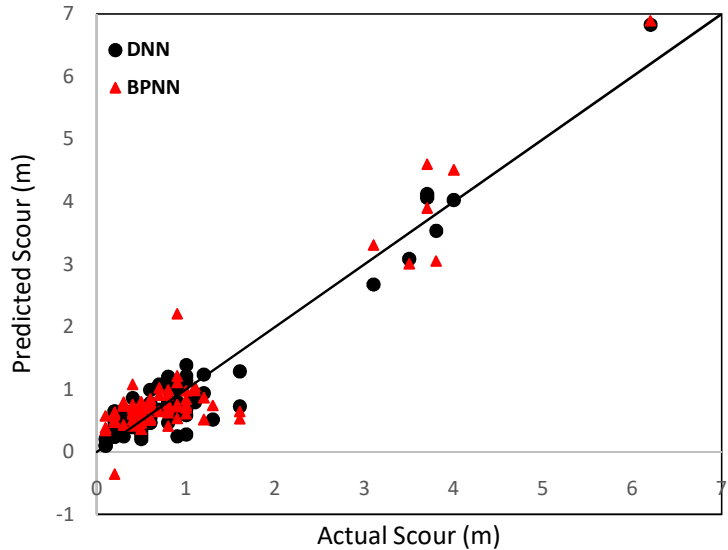

**Figure 1.** Actual vs predicted scour by DNN and BPNN using test data

Figure 2 represents the variation of actual and predicted scour depth with the number of test data using DNN and BPNN algorithm. It is evident from this figure that scour depth predicted by DNN is in good agreement with the actual scour depth. Availability of negative predicted value with BPNN may be considered as a drawback of this approach.

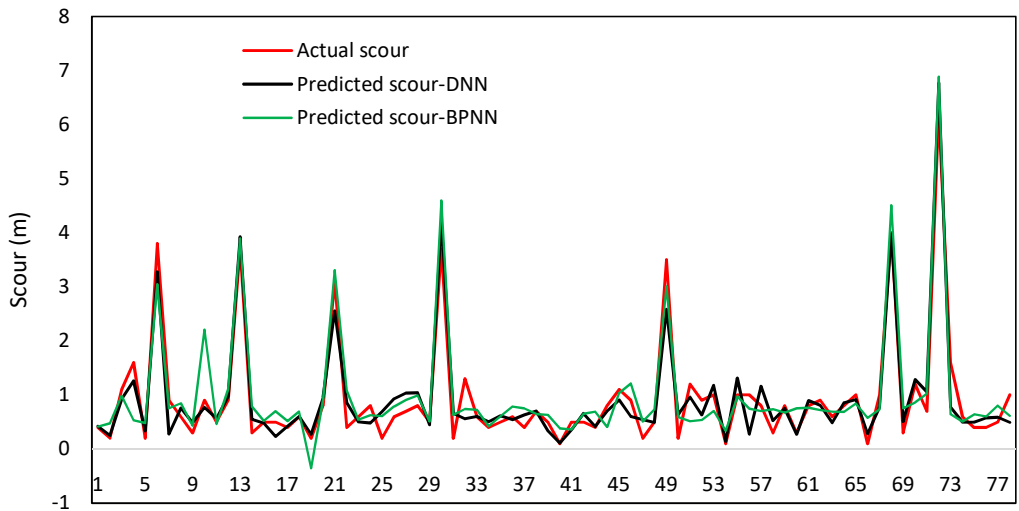

**Figure 2**. Actual and predicted scour using DNN and BPNN with test data set

## 5. CONCLUSIONS

This paper investigates the potential of deep neural network in predicting the local scour around bridge piers using field dataset. It can be concluded from this study is that deep neural networks are promising modeling approach and need to be further used for various problems related to water resource engineering to judge their full potential. Present study only explored the potential of dense layers (hidden layers) to create a deep neural network, it is planned in future to explore the potential of other hidden layers like long short-term memory (LSTM; Hochreiter and Schmidhuber 1997), choice of different updater, weight initialization methods and activation function.